\documentclass[11pt,a4paper]{article}

\usepackage{float}
\usepackage{microtype}
\usepackage{graphicx}

\usepackage{emnlp2020}
\aclfinalcopy % Uncomment this line for the final submission
\def\aclpaperid{12}%Enter the acl Paper ID here

\usepackage[utf8]{inputenc} % allow utf-8 input
\usepackage[T1]{fontenc}    % use 8-bit T1 fonts
\usepackage{url}            % simple URL typesetting
\usepackage{booktabs}       % professional-quality tables
\usepackage{amsfonts}       % blackboard math symbols
\usepackage{nicefrac}       % compact symbols for 1/2, etc.
\usepackage{microtype}      % microtypography
\usepackage{times}
\usepackage{latexsym}

\title{Multiple Sclerosis Severity Classification From Clinical Text}

\author{Alister D'Costa$^{1,2*}$, Stefan Denkovski$^{1,3*}$, Michal Malyska$^{1*}$, \\ \textbf{Sae Young Moon}$^{1,3*}$, \textbf{Brandon Rufino}$^{1,4*}$, \\ \textbf{Zhen Yang}$^5$, \textbf{Taylor Killian}$^{1,6}$, \textbf{Marzyeh Ghassemi}$^{1,6}$ \\
  $^1$University of Toronto, $^2$Ontario Institute for Cancer Research,\\ $^3$Toronto Rehabilitation Institute, $^4$Bloorview Research Institute,\\ $^5$Unity Health Toronto, $^6$Vector Institute \\ 
  $^*$\small Equal Contribution \\
  \small \texttt{\{alister.dcosta, stefan.denkovski, michal.malyska, sally.moon, brandon.rufino,} \\ \small \texttt {t.killian, marzyeh.ghassemi\}@mail.utoronto.ca,} \small \texttt{zhen.yang@unityhealth.to}}

\begin{document}

\maketitle

\begin{abstract}
Multiple Sclerosis (MS) is a chronic, inflammatory and degenerative neurological disease, which is monitored by a specialist using the Expanded Disability Status Scale (EDSS) and recorded in unstructured text in the form of a neurology consult note. An EDSS measurement contains an overall ‘EDSS’ score and several functional subscores. Typically, expert knowledge is required to interpret consult notes and generate these scores. Previous approaches used limited context length Word2Vec embeddings and keyword searches to predict scores given a consult note, but often failed when scores were not explicitly stated. In this work, we present MS-BERT, the first publicly available transformer model trained on real clinical data other than MIMIC. Next, we present MSBC, a classifier that applies MS-BERT to generate embeddings and predict EDSS and functional subscores. Lastly, we explore combining MSBC with other models through the use of Snorkel to generate scores for unlabelled consult notes. MSBC achieves state-of-the-art performance on all metrics and prediction tasks and outperforms the models generated from the Snorkel ensemble. We improve Macro-F1 by 0.12 (to 0.88) for predicting EDSS and on average by 0.29 (to 0.63) for predicting functional subscores over previous Word2Vec CNN and rule-based approaches. 
\end{abstract}

\section{Introduction}

Recent advancements of deep learning models with electronic health records (EHR) have shown a great deal of success in many clinical applications \cite{shickel2017deep}, such as disease detection \cite{choi2016retain}, diagnostics \cite{choi2017using}, risk predictions \cite{Futoma2015} and patient subtyping \cite{che2017rnn, baytas2017patient}. However, when the data within the EHR is presented in the form of narrative, unstructured clinical notes, extensive work is required by a professional to diagnose and generate labels for a patient \cite{PRATT1973}.

The development of pre-trained language models, namely Bidirectional Encoder Representations from Transformers (BERT), have significantly improved natural language processing (NLP) tasks within the general language domain \cite{Devlin2018}. However, in specialized domains such as the clinical one, the vocabulary, syntax and semantics differ significantly from general language \cite{Liu2012} and thus pretraining a language model on domain-specific texts is critical to improving performance. This is supported by the observed increase in performance on domain-specific NLP tasks when pretraining a BERT model on domain-specific texts \cite{lee_yoon_kim_kim_kim_so_kang_2019, Peng2019, alsentzer-etal-2019-publicly, Beltagy2019SciBERT}. Take for example BlueBERT \cite{Peng2019}, which has been further pretrained on over 4 billion words from PubMed abstracts and 500 million words from MIMIC-III \cite{JohnsonPollardShenEtAl2016} and has been shown to outperform BERT on multilabel classification from the Hallmarks of Cancers corpus \cite{Peng2019}.

Domain-specific language models, such as BlueBERT, still face several challenges for clinical NLP tasks. First, clinical texts must be de-identified of sensitive information, with the replacement of key tokens reducing the model's ability to interpret the text \cite{meystre_2014}. Second, texts from a specific clinical application may contain unique sub-language that the model was not trained on, hindering the model's performance. Third, transformer models have a fixed context length of 512 tokens that is significantly shorter than the average length of clinical texts \cite{Devlin2018}. As a result of truncating the text to fit the context length, the model is unable to analyze the entire text and may miss important information. These are the challenges of applying existing BERT models to specific clinical NLP tasks, which we have addressed through our contributions applied to a multiple sclerosis (MS) dataset.
\newline
\newline \textbf{Our contributions are as follows:}
\newline
\newline \textbf{[1]} A publicly available BERT based model pre-trained on over 70,000 MS consult notes, which we call MS-BERT.
\newline
\newline \textbf{[2]} A comprehensive pipeline for target predictions that integrates MS-BERT into a classifier, which we call MSBC. We apply MSBC to two tasks: (I) prediction of EDSS and functional subscores from neurological consult notes of MS patients and (II) generation of labels for an unlabelled consult note cohort.
\newline
\newline \textbf{[3]} Methods for data de-identification that preserves contextual information, optimized for fixed-context length models.
\newline
\newline \textbf{[4]} A novel approach to generate encounter level embeddings for documents larger than the BERT context window.
\newline
\newline \textbf{[5]} Semi-supervised labelling pipeline using the Snorkel framework \cite{snorkel} that increased the training data available for EDSS prediction and provided a quantitative analysis of silver-labelling strategies on real clinical applications.

\section{Methods}
\label{sec:msbc}

\textbf{De-identification of clinical text}. The consult notes used in this study contained sensitive information such as patient's name, phone numbers, physician's name and address. We de-identified the data using a curated database of patient and doctor information and regular expression matching. We replaced identifying pieces of information with specific tokens that met the following criteria: (1) the token was within the current BERT vocabulary, (2) the token had a similar semantic meaning to the word it replaced, and (3) the token was not found in the original data set. For example, all last names were replaced with "Salamanca". In doing so, we aimed to limit the loss of contextual information that results from de-identification. We also overcame challenges with sub-optimal placeholder replacements often present in clinical datasets, like MIMIC-III \cite{JohnsonPollardShenEtAl2016}. As an example, MIMIC-III may replace a patient's last name with "[**LAST NAME PLACEHOLDER**]", which is tokenized by BERT into at least 7 tokens (one for each square bracket, one for each star and at least one for the place holder within the brackets). A list of our de-identification replacements can be found within the appendices (contribution [3]).

\textbf{MS-BERT}. We used the de-identified consult notes to pre-train a language model optimized for NLP tasks related to MS, namely MS-BERT. MS-BERT is a BERT model that uses BlueBERT \cite{Peng2019} as its starting point, where BlueBERT is a BERT model pre-trained on PubMed abstracts \& MIMIC III note cohorts \cite{JohnsonPollardShenEtAl2016}. We used a masked language modeling (MLM) pre-training task \cite{Devlin2018} over all de-identified consult notes. The task used the bi-directional nature of the BERT model to predict a series of randomly selected masked tokens in a piece of text, allowing the model to learn the contextual meaning of the words in a sentence. This resulted in a language model that is optimized for understanding MS consult notes. The MS-BERT language model has been made available for use and is publicly accessible. The pretrained MS-BERT model can be found \href{https://huggingface.co/NLP4H/ms\_bert}{\underline{here}} (contribution [1]). %

\textbf{Encounter Level Embedding}. We generated encounter level embeddings for each consult note to address issues related to the limited context length of transformer models. Most transformer models have a context length limited to a number of sub-word tokens (512 in case of BERT \cite{Devlin2018}); however, the consult notes are often significantly longer. We separated consult notes longer than the context length into chunks of the maximum context length (in our case the length was 512 tokens). We then used MS-BERT to embed each chunk, resulting in a variable length output sequence of 768 dimensional vectors. 

\begin{figure*}[htp]
  \centering
  \includegraphics[width=15cm]{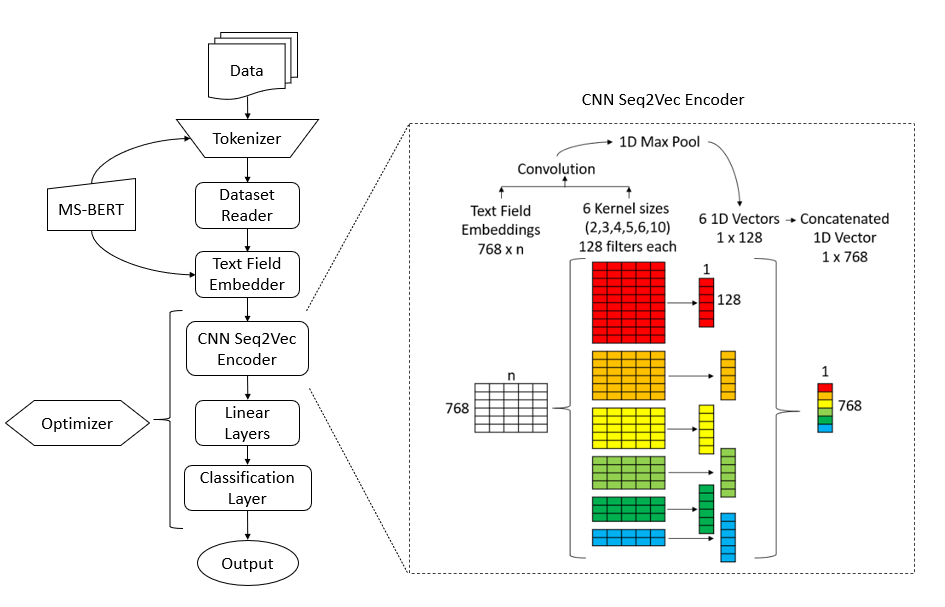}
  \caption{The MSBC architecture. We used a CNN described by \citet{CNNEncoder} to generate encounter level embeddings.}
  \label{figure:MSBERT_Model}
\end{figure*}

We explored 3 methods of converting the sequence of chunk level embeddings into a singular encounter level embedding: (1) taking the average across the sequence; (2) taking the max across the sequence; and (3) using a convolutional neural network (CNN) encoder based on \citet{CNNEncoder} included in the AllenNLP library. For more details see Figure \ref{figure:MSBERT_Model}.

In preliminary testing, the first two options under-performed the CNN encoder by a large margin ($\sim$60\%), thus we proceeded with the third option. Our final CNN encoder consists of six 1D convolutions with kernels of size [2, 3, 4, 5, 6, 10] and 128 filters each for a total of 768 dimensions in the output. This output is our final note embedding. We compared these full-length encounter level embeddings to embeddings that were generated using only a single context window (i.e. 512 tokens) and found that encounter level summaries were critical to model performance. 

% UPDATE
\footnote{Code for our pipeline and experiments are available \href{https://github.com/NLP4H/MSBC}{\underline{here}} (contributions [2,4,5])}
\textbf{MSBC}. Finally, we developed a custom classifier named MSBC (Multiple Sclerosis BERT CNN) to predict MS severity labels (EDSS or a functional subscore) using MS-BERT. MSBC is built using the AllenNLP \cite{Gardner2017AllenNLP} framework. A breakdown of MSBC is as follows. MSBC first reads in a consult note, tokenizes the text using the BERT vocabulary and then splits the tokens into chunks of size 512. MS-BERT weights are applied to each token chunk and all chunks for a note are then passed into the CNN based sequence to vector (Seq2Vec) encoder described above to pool the chunks and generate an encounter level embedding (i.e. a 1D vector of 768). This encounter level embedding is passed through 2 linear feed forward layers, acting as a dimension reduction step, before finally being passed to a linear classification layer to predict a label for the note. Figure \ref{figure:MSBERT_Model} shows an overview of MSBC's architecture. 

We trained and optimized MSBC for variables of interest, namely EDSS and functional subscores. Each note in the training set was passed through MSBC as described above. The resulting label was compared to the target label and a loss was computed. We used an AdamW optimizer to propagate errors back through the model, with a learning rate of 0.0005, weight decay of 0.01 and bias correction on a binary cross entropy loss function. We treat this as a classification problem instead of regression because EDSS is not uniform i.e. the difference between 3 and 4 is not the same as 4 and 5. We trained each model over 50 epochs using a batch size of 5 with 4 gradient accumulation steps. The model was saved at the end of each epoch if it had the best value for the validation metric. If during training the best validation metric was not beaten within 5 epochs, the trainer stopped early. A model for each prediction task was generated using MSBC and the train and validation sets described above. Once trained, we evaluated performance on the held out test set.

\textbf{Semi-Supervised Labelling}. Due to the costs of manually reviewing and labelling clinical texts, a significant majority of clinical texts in EHRs remain unlabelled \cite{garla_taylor_brandt_2013}. To leverage the full potential of all clinical text available and generate pseudo-labels for unlabelled data, we explored semi-supervised labelling using the Snorkel framework (v 0.9.3) \cite{snorkel}. Snorkel facilitates weak supervision of unlabelled data given weak heuristics and classifiers (i.e. labelling functions or LFs) \cite{Ratner2016,snorkel}. Snorkel's Label Model, a generative model, combines the predictions and generates a single confidence weighted label per data point. Snorkel does this by using the LFs' observed agreement and disagreement rates to estimate the unknown accuracy of the LF's. Snorkel then learns and models the accuracies of the LFs to combine the labels and generate the final label per data point \cite{ratner_Snorkel_2}. To identify the optimal combination of LFs to label the unlabelled notes, we evaluated the performance of task predictions on various Snorkel ensembles. The model that yielded the highest performance on our validation-set was chosen to be used to label the unlabelled notes. 

We created two additional models using the MSBC architecture: MSBC$+$, trained on a combination of labelled and pseudo-labelled data and MSBC-silver, which is a model trained on only pseudo-labelled data. We pursued the development of MSBC-silver as an attempt to see if we could reconstruct our model without access to the original labelled data, similarly to \citet{Krishna2020Thieves}.

\section{Experiments}

Multiple sclerosis (MS) is one of the most common non-traumatic disabling neurological condition among young adults worldwide \cite{Ploughman2014,Wade2014}. Onset of MS typically occurs between the ages of 20 to 40 years, with women more often affected than men \cite{Ploughman2014}. MS is a disease that impacts the central nervous system (CNS) \cite{Goldenberg2012}, leading to the degradation of myelin sheathing and axons within the nervous system. This degradation is highly varied and unpredictable in both location and intensity within the body. Resulting symptoms include but are not limited to: visual impairment, loss of balance, numbness, bladder dysfunction and fatigue \cite{Calabresi2004}. 

MS is typically monitored by the Expanded Disability Status Scale (EDSS) \cite{Kurtzke1983}. EDSS is used to evaluate the degree of CNS impairment on a scale from 0 to 10. EDSS also includes eight functional subscores \cite{Kurtzke1983} such as an ambulation score and a visual score. A full list of functional subscores is found within Table \ref{table:subscore_results} and their respective descriptions can be found in the appendices.

EDSS and functional subscores are discussed in a patient's consult note, dictated by a physician and manually transcribed. EDSS is determined by a combination of functional subscores and is typically stated within consult notes. However, functional subscores are not typically stated within a consult note and need to be derived from contextual information about the patient's health. Traditionally, both EDSS and functional subscores are manually derived by an expert within the field and logged into the patient's health record. Minute differences in patient descriptions can correspond to different EDSS and functional subscore values. Through consultation with MS healthcare professionals we expect the qualitative descriptions of MS symptoms contained within the clinical notes to remain uniform across healthcare systems. 

\begin{figure*}[htp]
  \centering
  \includegraphics[width=12.7cm]{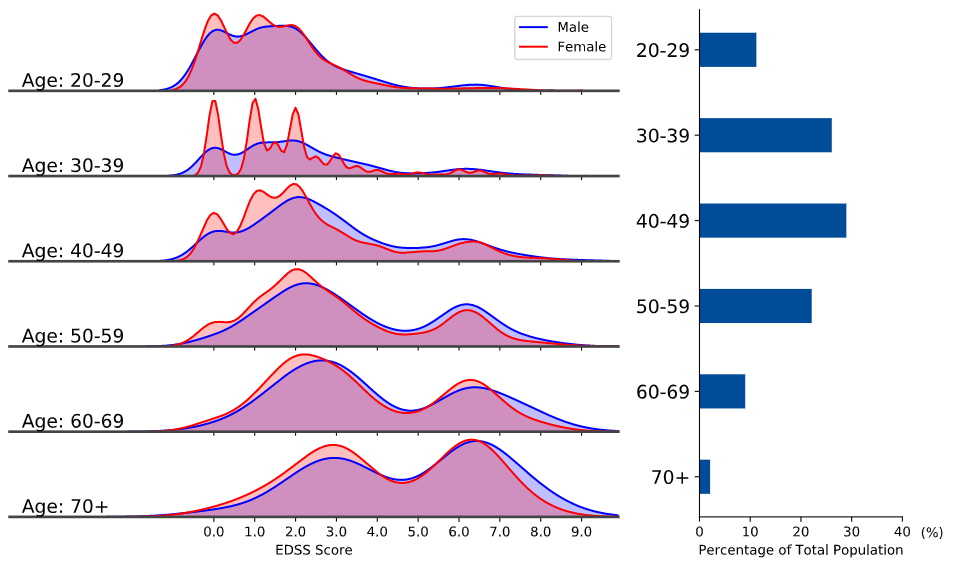}
  \caption{Distribution of EDSS scores varied by age and gender.}
  \label{figure:edss_age_gender}
\end{figure*}

\subsection{Data}
\label{sec:data}

The dataset, compiled by a leading MS research hospital, contains approximately 70,000 MS consult notes for about 5,000 patients, totaling over 35.7 millon words. These notes were collected from patients who visited this hospital's MS clinic between 2015 to 2019. Of the 70,000 notes approximately 16,000 are manually labeled by a research assistant for EDSS and functional subscores. The gender split within the dataset was observed to be 72\% female and 28\% male as shown in Figure \ref{figure:edss_age_gender} and reflecting the natural discrepancy in MS \cite{Harbo2013}.

Once de-identified, data was separated into labelled and unlabelled sets. The labelled set was further separated into test ($\sim$30\%), train ($\sim$50\%) and validation ($\sim$20\%) subsets. When designing the splits for our data, we wanted to ensure that we could accurately predict EDSS and functional subscores on new notes for both current and new patients and to reduce any gender bias that may occur from population discrepancy. First we stratified by gender. Then we either fully contained the notes of one patient within a subset or divided the patients notes across subsets chronologically. This allowed for earlier notes to be used for training, and later notes for validation and test. Due to de-identification of notes the risk of information leakage between subsets is minimized. 

\subsection{Experiment 1: EDSS and Functional Subscore Prediction}
\label{sec:task1}

\textbf{Previous Work}. Previous approaches to extract information from MS consult notes have typically relied on keyword searches \cite{Davis2015,Damotte2019}. We refer to the collection of these searches as the rule-based (RB) approach. Word2Vec embeddings used with a convolutional neural network (CNN), have been shown to be successful in clinical tasks such as creating explainable predictions of medical codes from clinical text \cite{Mullenbach2018}. Previous work done at our affiliated MS hospital used Word2Vec embeddings and a CNN model to generate EDSS predictions. Best results were achieved by incorporating the RB approach with the Word2Vec CNN. This method first used the RB approach to extract keywords and phrases that infer EDSS scores. If the RB approach was unable to predict a score, then the prediction from the Word2Vec CNN model was used. More information on the development of the CNN model can be found in the appendices. In this work, we compared the performance on predicting EDSS and functional subscores between the: (1) Word2Vec CNN, (2) a sequential approach using RB plus Word2Vec CNN, (3) MSBC, and (4) a sequential approach using RB plus MSBC.

Additional baselines were established with term frequency-inverse document frequency (tf-idf) features. These features have been successful in various clinical NLP tasks \cite{Bhattarai2009,NarayanShukla2020,Boag2018}. A number of baseline models were developed on top tf-idf features such as: support vector machines (SVM), logistic regression (LR) and linear discriminant analysis (LDA). Due to a lack of performance on the easier task of predicting EDSS scores (see Table \ref{table:edss_results}), they were not evaluated for the prediction of functional subscores.

\textbf{Results}. Our results for EDSS prediction are summarized in Table \ref{table:edss_results} and functional scores in Table \ref{table:subscore_results}. MSBC achieves top performance in both tasks in all metrics. For EDSS prediction, Macro-F1 and Micro-F1 are improved upon by 0.11 and 0.043 respectively. For functional subscore prediction, we see a significant improvement of over 0.35 in Macro-F1 and almost 0.15 in Micro-F1.

\begin{table*}[!htbp]
\centering
\caption{EDSS prediction performance for all models. Higher values indicate stronger performance and highest values are bolded.}
\label{table:edss_results}
\resizebox{0.5\textwidth}{!}{\begin{tabular}{lll} 
\toprule
Model                              & Macro-F1          & Micro-F1           \\ 
\midrule
Multiple Sclerosis Bert Classifier (MSBC)                          & \textbf{0.88296}  & \textbf{0.94177}   \\
MSBC Truncated (only first 512 tokens)    &0.74680 &0.90086   \\
Rule-Based (RB) + Word2Vec CNN                                  & 0.76817          & 0.89668          \\
\midrule
RB + MSBC                                 & 0.86625          & 0.92987          \\
Word2Vec CNN                       & 0.66475           & 0.88144            \\
RB                                 & 0.76694          & 0.83761          \\
BlueBERT CNN                       & 0.51000           & 0.81000            \\
Linear SVC                         & 0.48503           & 0.74452            \\
LDA                                & 0.50122           & 0.74390            \\
SVC RBF                            & 0.45877           & 0.72428            \\
Log Reg                    & 0.45763           & 0.71175            \\
\bottomrule
\end{tabular}}
\end{table*}

\begin{table*}[!ht]
\centering
\caption{Sub-score prediction performance differences between baseline and MSBC. Higher values indicate stronger performance. Highest values are bolded. It should be noted that low to no support for the highest levels of sub-scores impacted Macro-F1.}

\label{table:subscore_results}
\begin{tabular}{l|cc|cc|cc} 
\toprule
\multicolumn{1}{l}{Models} & \multicolumn{2}{c}{ \textbf{MSBC} } & \multicolumn{2}{c}{ \textbf{RB}} & \multicolumn{2}{c}{\textbf{RB + Word2Vec}}  \\ 
\midrule
Subscore                  & Macro-F1        & Micro-F1               & Macro-F1 & Micro-F1              & Macro-F1 & Micro-F1                         \\ 
\midrule
Ambulation                & 0.6980          & 0.88797                & 0.2710   & 0.5627                & 0.2674   & 0.5155                         \\
Bowel Bladder             & 0.6039          & 0.86619                & 0.2773   & 0.5525                & 0.2027  & 0.5209                           \\
Brain Stem                & 0.5842          & 0.90356                & 0.4174   & 0.5694                & 0.3712   & 0.6598                           \\
Cerebellar                & 0.6437          & 0.85707                & 0.4927   & 0.6120                & 0.4188   & 0.5908                           \\
Mental                    & 0.5496          & 0.79470                & 0.3643   & 0.5586                & 0.3003   & 0.5499                           \\
Pyramidal                 & 0.7192          & 0.87755                & 0.4173   & 0.5128                & 0.4028   & 0.5598                           \\
Sensory                   & 0.5570          & 0.87518                & 0.4082   & 0.4173                & 0.3485   & 0.5603                           \\
Visual                    & 0.7153          & 0.93855                & 0.5020   & 0.4082                & 0.4207   & 0.6986                           \\ 
\midrule
 \textbf{Mean}            & \textbf{0.6339} & \textbf{0.8751 }       & 0.3937   & 0.5737                & 0.3416 & 0.5820                           \\
\bottomrule
\end{tabular}
\end{table*}

\textbf{Discussion}. The significant improvement of MSBC, especially in Macro-F1, indicates that MS-BERT is better able to distinguish nuances within text that characterize different EDSS and functional subscores. Interestingly, the Word2Vec CNN outperformed BlueBERT, which is likely attributed to the fact that Word2Vec was pre-trained on our corpus of text. Also, our different method of de-identifying data from MIMIC-III (which BlueBERT was pre-trained on), may have reduced BlueBERT's effectiveness. However, the contextually similar token replacement should limit this impact.

We see a strong improvement in functional subscore predictions over the baselines. While EDSS is stated directly in notes, functional subscores are typically referenced indirectly. This makes it more difficult for a rule based approach and simple models to learn the contextual information required to assess scores. Furthermore, EDSS and functional subscores also suffer from a high level of disagreement among clinicians, particularly for the sensory and mental categories \cite{PiriCinar2018}. The level of disagreement typical is lower for EDSS scores greater than 5.5 and in general does not exceed 1. At two clinics, examined EDSS scores differed by 0.5 for up to 29\% of patients and by 1 for up to 50\% of patients. This level of subjectivity and variability within the true labels may make it difficult for the model to predict accurately. That said, due to the contextual awareness brought by MS-BERT, MSBC shows strong improvement from previous work when predicting functional subscores. Additionally, the labels for functional subscores were generated post-examination by trained clinicians based on the contents of notes. Therefore, missing information from notes led to missing labels for certain functional subscores, resulting in varying levels of support for different scores. 

MSBC under-performed on classes with low support. The bottom 25\% of classes in terms of support averaged an F1 score of 0.78, which was 0.1 lower than the mean for all classes. However, classes with low support are typical of EDSS due to its bi-modal distribution \cite{Meyer-Moock2014}. This is a result of the non-linear method of determining EDSS based on certain heuristics and conditions (i.e. the difference between an EDSS score of 3 to 4, is not the same as 4 to 5). 

To help understand why and when rule based approaches failed, we looked at performance of the models only on notes that rule based approaches were not able to label EDSS scores (see appendices). This accounted for around 12\% of the notes and we see very poor performance for all other models with F1 scores below 0.36 (and very high F1-scores for those rule based were able to label), while MSBC is still able to achieve an F1 score above 0.6. This may indicate that a certain portion of notes that contain poor quality information and may be ``trickier" to label. These ``tricky" notes could be notes that state ``no change" or ``similar" results to past notes, without restating those scores for example. However, it is predicted that MSBC was still able to outperform other models through its ability to understand contextual information embedded in the text.

\subsection{Experiment 2: Semi-Supervised Labelling of EDSS}

We evaluated the effectiveness of the Snorkel ensembles and compared the performance of: (1) MSBC (which has been observed in Experiment 1), (2) MSBC$+$, and (3) MSBC-silver.

We hereon refer to two types of labels: (1) gold labels (n$\sim$16,000), which were manually obtained by a professional at our MS clinic and are considered truth in our experiments, and (2) silver labels (n$\sim$54,000), which were generated from the model chosen for EDSS labelling.

\textbf{Results}. Various Snorkel ensembles were evaluated as presented in Table \ref{table:snorkel_results}. Only the LF combinations that included MSBC were evaluated as MSBC had the best EDSS prediction performance. From the F1 scores, we observe that MSBC alone outperforms all ensembles that contain MSBC by at least 0.02 on Macro-F1. The addition of weaker classifiers consistently decreased the ensemble's performance. Furthermore, we observe that the amount of conflict for MSBC (i.e. fraction of data MSBC disagrees with for at least one other LF) increases as weaker classifiers are added to the ensemble. 

\begin{table*}
\centering
\caption{EDSS predictions results for Snorkel ensembles containing MSBC. Conflicts reflect the fraction of data that MSBC disagrees with at least one other LF. Highest values are bolded.}
\label{table:snorkel_results}
\begin{tabular}{llll} 
\toprule
 \textbf{Ensemble combinations}              & Macro-F1 & Micro-F1 & Conflicts  \\ 
\hline
MSBC                                    & \textbf{0.88296}  & \textbf{0.94177}  & N/A        \\
MSBC + Rule Based LFs (RB LFs)                      & 0.86617  & 0.93363  & 0.23471    \\
MSBC + RB LFs + Word2Vec           & 0.78582  & 0.91901  & 0.33229    \\
MSBC + RB LFs + Word2Vec + LDA     & 0.77004  & 0.88917  & 0.46796    \\
MSBC + RB LFs + Word2Vec + TFIDFs & 0.55728  & 0.82592  & \textbf{0.55145}    \\
\bottomrule
\end{tabular}
\end{table*}

\begin{table*}
\centering
\caption{Performance of MSBC predicting EDSS using different label types. Gold labels (n=16,000) were manually obtained by a professional at our MS clinic and are considered truth in our experiments. Silver labels (n=54,000) were generated from MSBC predictions which was trained on gold labels. Higher values indicate stronger performance. Highest values are bolded.}
\label{table:ssl}
\begin{tabular}{llll} 
\toprule
Model   & Trained on                       & Macro-F1          & Micro-F1           \\ 
\midrule
MSBC  & Gold Labels                        & \textbf{0.88296}  & \textbf{0.94177}   \\
MSBC$+$ & Silver + Gold Labels                   & 0.86238           & 0.92569            \\
MSBC-silver        & Silver Labels              & 0.82922           & 0.91442            \\
\bottomrule
\end{tabular}
\end{table*}
From the above analysis, we concluded that MSBC alone, out of all Snorkel ensembles, performs the best and therefore was chosen to generate silver-labels for the unlabelled neurology notes. Various models were trained using the MSBC architecture and are presented in Table \ref{table:ssl}. The best version of MSBC was the model trained solely on gold label data (our original MSBC). Macro-F1 score and Micro-F1 score are observed to drop in MSBC$+$. MSBC-silver was the worst out of the 3 variations with a Macro-F1 of 0.83 and Micro-F1 of 0.91 but is still observed to outperform the previous best baseline (RB+Word2Vec CNN presented in Table \ref{table:edss_results}) by an approximate Macro and Micro-F1 of 0.06 and 0.02 respectively.

\textbf{Discussion}. MSBC alone performs better than all Snorkel ensembles. The performance of the ensembles consistently decreased as more weak classifiers and heuristics were added. We hypothesize that the drop in performance is due to the fact that the Snorkel's Label Model learns to predict the accuracy of the LFs based on observed agreements and disagreements. It also assumes conditional independence among the LFs \cite{ratner_Snorkel_2}. This result is not surprising given that the qualitative analysis of errors showed that MSBC was almost strictly an improvement over the Rule-Based approach. MSBC only struggled with notes that had EDSS indicated in the roman numeral `iv' (which could be misconstrued to be the lower-case acronym for intravenous) and notes where patient complaints of their symptoms were contained in a different note chunk than the physician findings which contradicted those symptoms. In all other cases, the model made no significant (off by no more than 0.5-1 on the EDSS scale) errors compared to the weak heuristics. Therefore in the presence of a strong LF, such as MSBC, we suspect that the addition of weaker LFs introduce disagreements with MSBC and thus decreased predictive performance. Furthermore, all LFs were developed based on the same labelled training data (for example, tf-idf models were trained on the same training set). Hence, it is likely that the LFs were correlated, which violated the conditional independence assumption made by Snorkel and compromised prediction accuracy.  

Our model trained on silver labeled data, MSBC-silver, performs worse than MSBC by 0.03-0.06. This small decrease in performance indicates that our model is able to relearn its own distribution and helps validate its performance. MSBC-silver outperformed all previous baselines on the EDSS prediction task. The strong results of MSBC-silver helps show the effectiveness of using MSBC as a labelling function. This work shows potential to reduce tedious hours required by a professional to read through a patient's consult note and manually generate an EDSS score.

\section{Concluding Thoughts}

In this work we present methods to overcome the challenges that arise when applying a modern transformer model on a specific clinical NLP task, specifically MS severity prediction. We did this through: (1) de-identifying clinical texts in a way that preserves contextual meaning; (2) generating encounter level embeddings to eliminate loss of information resulting from the limited context length of transformer models; (3) further pretraining a BERT model on MS consult notes to build a language model (MS-BERT) with better understanding of MS clinical notes; (4) developing a classifier (MSBC) that uses MS-BERT to achieve state of the art performance on predicting EDSS and functional subscores; and (5) using our classifier to generate labels for previously unlabelled data, showing its effectiveness as a labelling model.

We believe that the MS-BERT language model and its improved ability to understand MS consult notes will aid clinicians in the diagnosis and treatment of MS. Furthermore, we believe that being trained on more clinical text, MS-BERT has the potential to improve other NLP tasks within the clinical domain. 

% We set out to predict EDSS and functional subscores using MSBC. We first presented MS-BERT, a language model specifically pre-trained on MS consult notes. We next showed that we could overcome BERT's limited context length by using a CNN to effectively combine numerous chunk level embeddings into an encounter level embedding. Through the use of the MS-BERT model we were able to predict EDSS and functional subscores with better performance than previously reported baselines, providing a model that can effectively be used in a clinical setting. This model was then used to generate labels for previously unlabelled data, showing its effectiveness as a labelling model.

% \subsection{Limitations and Potential Improvements}

% Our de-identification procedure was based on a set of regex match rules that ended up erasing some terminology and inserting our token replacement for surname into the notes (ie. people's surnames are sometimes the same as common words). Instead we could try a Named Entity Recognition model (NER) and manually ensure that all regex matches left were those of common nouns.

% Our de-identification should have replaced doctor names with different tokens than patient names to not lose out on context.

% Consider using embeddings for each token and apply our sequence combining CNN model instead of just combining [CLS] token embeddings to represent chunks.

\subsection{Future Work}

First, we are in the process of implementing an interpretability module that would provide per-word attentions instead of the per-sub-word-token attentions available out-of-the-box. Second, we want to evaluate MS-BERT's performance on other language tasks such as relation extraction, sentence similarity, inference tasks, and question answering within the clinical space. Third, we would like to experiment with other note-level embeddings and model architectures, such as the CNN presented by Kim 2014 \cite{Kim2014}. While we are pleased with the performance of MSBC, we would like to demonstrate that our approach (the methods for de-identifying data, fine-tuning a language model, the generation of encounter level embeddings and our custom classifier) can be applied on other clinical datasets. Also, we would like to pre-train longer context transformer models such as the Reformer \cite{Kitaev2020Reformer} which targets longer context windows and compare it to Clinical BERT which is tailored for the clinical domain \cite{alsentzer-etal-2019-publicly}. Finally, we would like to see if using token level embeddings as inputs to our CNN encoder, along with replacing some tokens with more clinically relevant ones in the base BERT vocabulary could improve encounter level embedding quality.
\\

\section{Acknowledgments}
We would like to thank the researchers and staff at the Data Science and Advanced Analytics (DSAA) team at St. Michael’s Hospital, for providing consistent support and guidance throughout this project. We would also like to thank Dr. Marzyeh Ghassemi, and Taylor Killan for providing us the opportunity to work on this exciting project. Lastly, we would like to thank Dr. Tony Antoniou and Dr. Jiwon Oh from the MS clinic at St. Michael’s Hospital for their support on the neurological examination notes.

\label{app:theorem}

%\section*{References}
\bibliographystyle{acl_natbib}
\bibliography{references}

\newpage
\appendix
\onecolumn
\section*{A De-identification of Clinical Text}
\label{sec:appendix}

\begin{table}[htp]
\centering
\caption{Full breakdown of word and category replacements for note de-identification.}
\label{table:de_id_dictionary}
\begin{tabular}{ll}
\toprule
\textbf{Value}            & \textbf{Replacement} \\ 
\midrule
Last / Family Names       & Salamanca            \\ \midrule
Female First Names        & Lucie                \\ \midrule
Male First Names          & Ezekiel              \\ \midrule
Phone/Fax                 & 1718                 \\ \midrule
MRN/PID                   & 999                  \\ \midrule
Dates / DOB               & 2010s                \\ \midrule
Time                      & 1610                 \\ \midrule
Addresses                 & Silesia              \\ \midrule
Location/Hospital/Clinics & Troy                 \\ \bottomrule
\end{tabular}
\end{table}

\section*{B Functional Subscores for EDSS}

\begin{table*}[htp]
\centering
\caption{Functional subscores for EDSS.}
\label{table:subscores}
\begin{tabular}{ll}
\toprule
\textbf{Functional Scores}            & \textbf{Description} \\ 
\midrule
Visual Function       & Ability to read of eye chart at 20 feet            \\ \midrule
Brainstems        & Eye movement, balance, hearing, numbness, swallowing, speech                \\ \midrule
Pyramidal          & Reflexes, limb strength, motor performance             \\ \midrule
Cerebellar                & Muscle coordination and control (ataxia)                \\ \midrule
Sensory                  & Ability to detect light touch or vibration                  \\ \midrule
Bowl and Bladder              & Control and correct function of bladder and bowl functions                \\ \midrule
Cerebral                   & Depression, mental alertness (mentation)                 \\ \midrule
Ambulation                 & Ability to walk unimpaired              \\ \bottomrule
\end{tabular}
\end{table*}

\section*{C Baseline Models}
\textbf{Term Frequency-Inverse Document Frequency}
\newline
We trained a number of baseline models on top of our tf-idf features, finding that our max feature space was optimal at 1500 tokens. After hyper-parameter tuning our tf-idf baseline models, we observed that the following performed best for predicting EDSS scores:

\begin{itemize}
  \item Support vector classification (SVC) with tuned regularization parameter ‘C’ equal to 1. Both linear and radial basis function (RBF) kernels were generated based on their strong performance in this classification task.
  \item Linear discriminate analysis (LDA) with a singular value decomposition solver.
  \item Logistic regression (LR) using a limited-memory BFGS (lbfgs) solver with ‘l2’ regularization and inverse regularization strength, ‘C’, equal to 100. This model also considered class weights within the training set.
\end{itemize}

\textbf{Word2Vec and Convolution Neural Networks}
\newline
Word2Vec models \cite{Mikolov2013,Choi2016} take a corpus of text and learn vector representations, called embeddings, for each word \cite{Che2017}. Words with similar context have been observed to have close embeddings in the vector space. 

CNNs have been observed to work well in a variety of clinical tasks. For example, CNN architectures have proved successful in relation extraction \cite{Sahu2016}, risk prediction \cite{Che2017}, the extraction of medical events from clinical notes \cite{Li2016}, and clinical named entity recognition \cite{Wu2017}.

Previous work done at our collaborating hospital used a 200-dimensional Word2Vec embedding trained on all MS consult notes (n=75,009) with a window size of 10 and a minimum count of 2. Next, they converted all tokenized notes into their word vector representations. While doing so, they set a maximum note length of 1,000 tokens and zero padded notes as necessary. They then designed a 3-dimensional input sequence (batch size x 1000 x 200). This input sequence was fed into a Keras \cite{chollet2015keras} implementation of the CNN architecture described by Kim 2014 \cite{Kim2014}. Finally, using convolutional layers (with max pooling), and fully connected layers (with softmax output), they trained their CNN model using the RMSProp optimizer with early stopping.

\section*{D Detailed Breakdown of MSBC prediction on EDSS}
\begin{table*}[htp]
\centering
\caption{Performance of MSBC across all values for EDSS.}
\label{table:msbc_breakdown}
\begin{tabular}{lllll}
\toprule
\multicolumn{5}{c}{\textbf{MSBC's Breakdown}}                  \\ \midrule
EDSS                  & Precision & Recall & F1     & Support \\ \midrule
0                      & 0.9764    & 0.9805 & 0.9784 & 717     \\ \midrule
1.0                      & 0.9605    & 0.9679 & 0.9642 & 779     \\ \midrule
1.5                      & 0.9751    & 0.9333 & 0.9534 & 420     \\ \midrule
2.0                      & 0.9365    & 0.9708 & 0.9533 & 926     \\ \midrule
2.5                      & 0.9410    & 0.9280 & 0.9344 & 361     \\ \midrule
3.0                      & 0.9413    & 0.9436 & 0.9425 & 408     \\ \midrule
3.5                      & 0.9362    & 0.8980 & 0.9167 & 196     \\ \midrule
4.0                      & 0.9632    & 0.9562 & 0.9597 & 137     \\ \midrule
4.5                      & 0.8605    & 0.7400 & 0.7957 & 50      \\ \midrule
5.0                      & 0.9157    & 0.8837 & 0.8994 & 86      \\ \midrule
5.5                     & 0.8889    & 0.8889 & 0.8889 & 81      \\ \midrule
6.0                     & 0.8689    & 0.9339 & 0.9002 & 227     \\ \midrule
6.5                     & 0.9247    & 0.8984 & 0.9113 & 246     \\ \midrule
7.0                     & 0.7761    & 0.7647 & 0.7704 & 68      \\ \midrule
7.5                     & 0.9286    & 0.6842 & 0.7879 & 38      \\ \midrule
8.0                     & 0.8889    & 0.8000 & 0.8421 & 30      \\ \midrule
8.5                     & 0.7500    & 0.9231 & 0.8276 & 13      \\ \midrule
9.0                     & 0.7143    & 0.6250 & 0.6667 & 8       \\ \midrule
\textbf{Mean}          & 0.8970    & 0.8734 & 0.8830 & 4791    \\ \midrule
\textbf{Weighted Mean} & 0.9420    & 0.9417 & 0.9414 & 4791    \\ \bottomrule
\end{tabular}
\end{table*}

\begin{figure*}[htp]
  \centering
  \includegraphics[width=8.5cm]{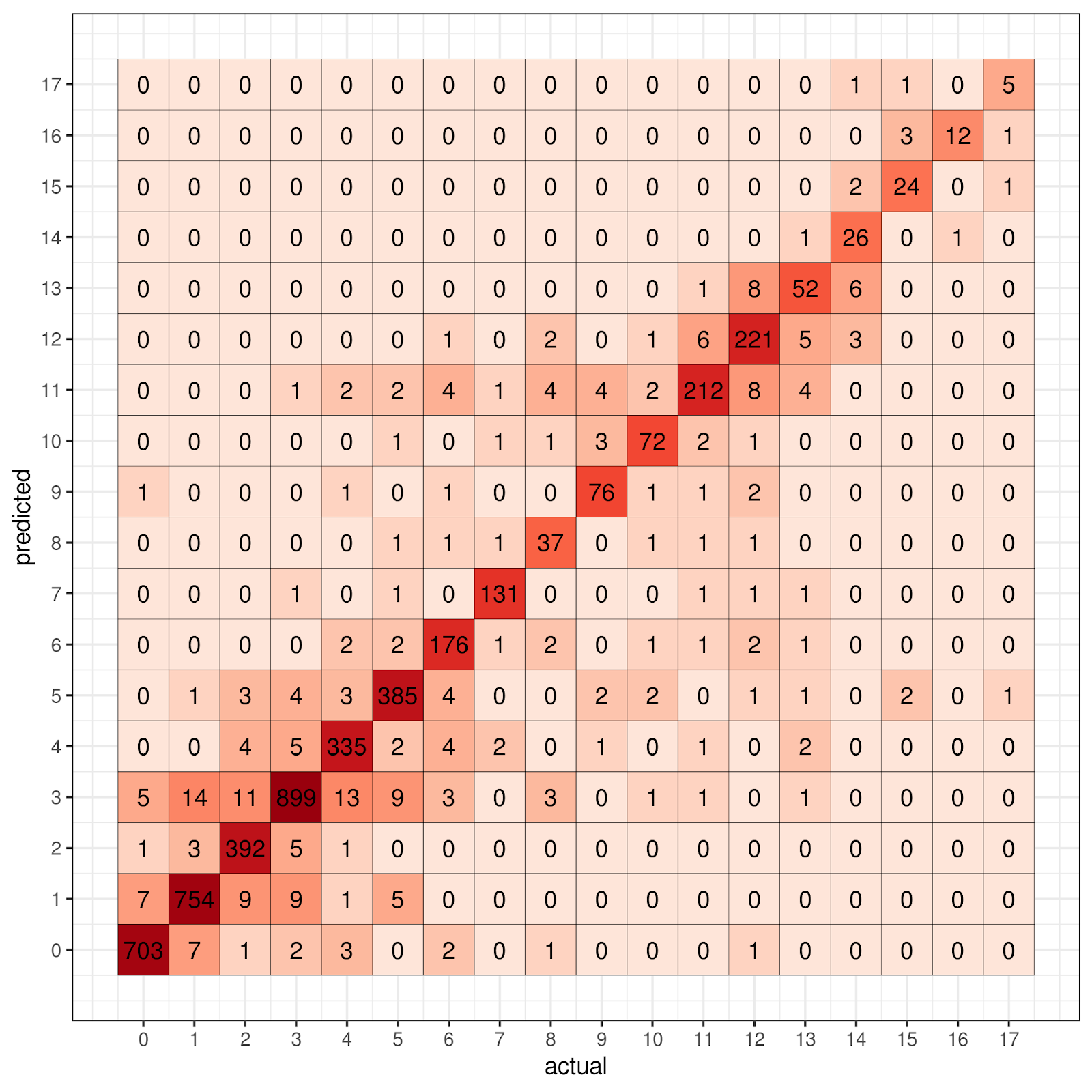}
  \caption{Heat map showing the distribution of predictions from our model compared to true values. Tight grouping is noticed in high levels of support, and less grouping where there is less support.}
  \label{figure:heatmap}
\end{figure*}

\section*{E Performance of MSBC on 'Tricky' Notes}
\begin{table*}[htp]
\caption{EDSS prediction across notes that were not found via a key word search. Bolded scores represent best model performance.}
\centering
\begin{tabular}{llll}
\toprule
\multicolumn{4}{c}{\textbf{EDSS Prediction on Samples that Rules were Unable to Label}} \\ \midrule
Model                           & Macro-F1         & Micro-F1         & Weighted-F1      \\ \midrule
MSBC                            & \textbf{0.49942} & \textbf{0.61268} & \textbf{0.60340} \\
RB + Word2Vec (Bench Mark)      & 0.19297          & 0.33275          & 0.32934          \\
Word2Vec CNN                    & 0.19297          & 0.33275          & 0.32934          \\
SVC RBF                         & 0.26748          & 0.40493          & 0.36611          \\
Log Reg Baseline                & 0.24783          & 0.35916          & 0.34876          \\
LDA                             & 0.23374          & 0.33627          & 0.32295          \\
Linear SVC                      & 0.18703          & 0.30634          & 0.29474          \\ 
\bottomrule

\end{tabular}
\end{table*}

\begin{table*}[htp]
\caption{EDSS prediction across notes that were found via a key word search. Bolded scores represent best model performance.}
\centering
\begin{tabular}{llll}
\\ \toprule
\multicolumn{4}{c}{\textbf{EDSS Predictions on Samples that Rules were Able to Label}}   \\ \midrule
Model                           & Macro-F1         & Micro-F1         & Weighted-F1      \\ \midrule
MSBC                            & \textbf{0.95363} & \textbf{0.98603} & \textbf{0.98599} \\
RB + Word2Vec CNN (Bench Mark)  & 0.93298          & 0.97253          & 0.97259          \\
Word2Vec CNN                    & 0.79170          & 0.95525          & 0.95393          \\
LDA                             & 0.53302          & 0.79872          & 0.80062          \\
Linear SVC                      & 0.52528          & 0.80346          & 0.80861          \\
SVC RBF                         & 0.48367          & 0.76723          & 0.75366          \\
Log Reg Baseline                & 0.48057          & 0.75918          & 0.75845 \\ \bottomrule   
\end{tabular}
\end{table*}

\newpage
\section*{F Exploratory Data Analysis}

\begin{figure*}[htp]
  \centering
  \includegraphics[width=10cm]{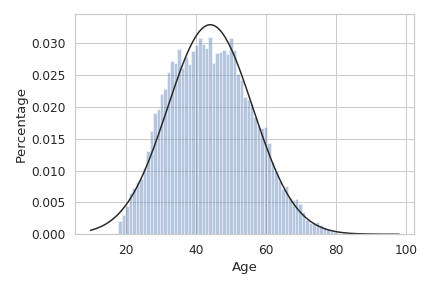}
  \caption{Distribution of age within the data set.}
  \label{figure:age_hist}
\end{figure*}

\begin{figure*}[htp]
  \centering
  \includegraphics[width=10cm]{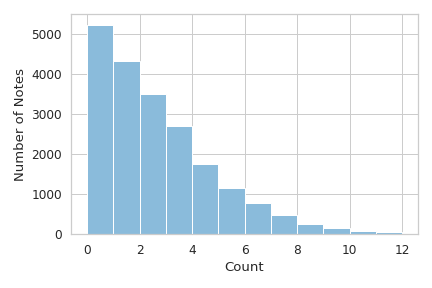}
  \caption{Histogram showing the number of notes per patient.}
  \label{figure:notes_per_pat_hist}
\end{figure*}

\begin{figure*}[htp]
  \centering
  \includegraphics[width=11cm]{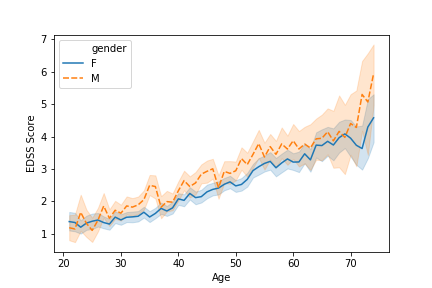}
  \caption{Plot of mean EDSS score vs age.}
  \label{figure:edss vs age (1)}
\end{figure*}

\begin{figure*}[htp]
  \centering
  \includegraphics[width=10cm]{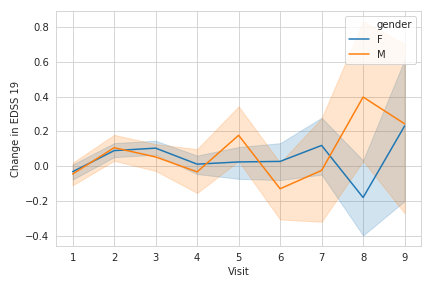}
  \caption{Change of EDSS score in subsequent visits.}
  \label{figure:edss vs age (2)}
\end{figure*}

\begin{figure*}[htp]
  \centering
  \includegraphics[width=15cm]{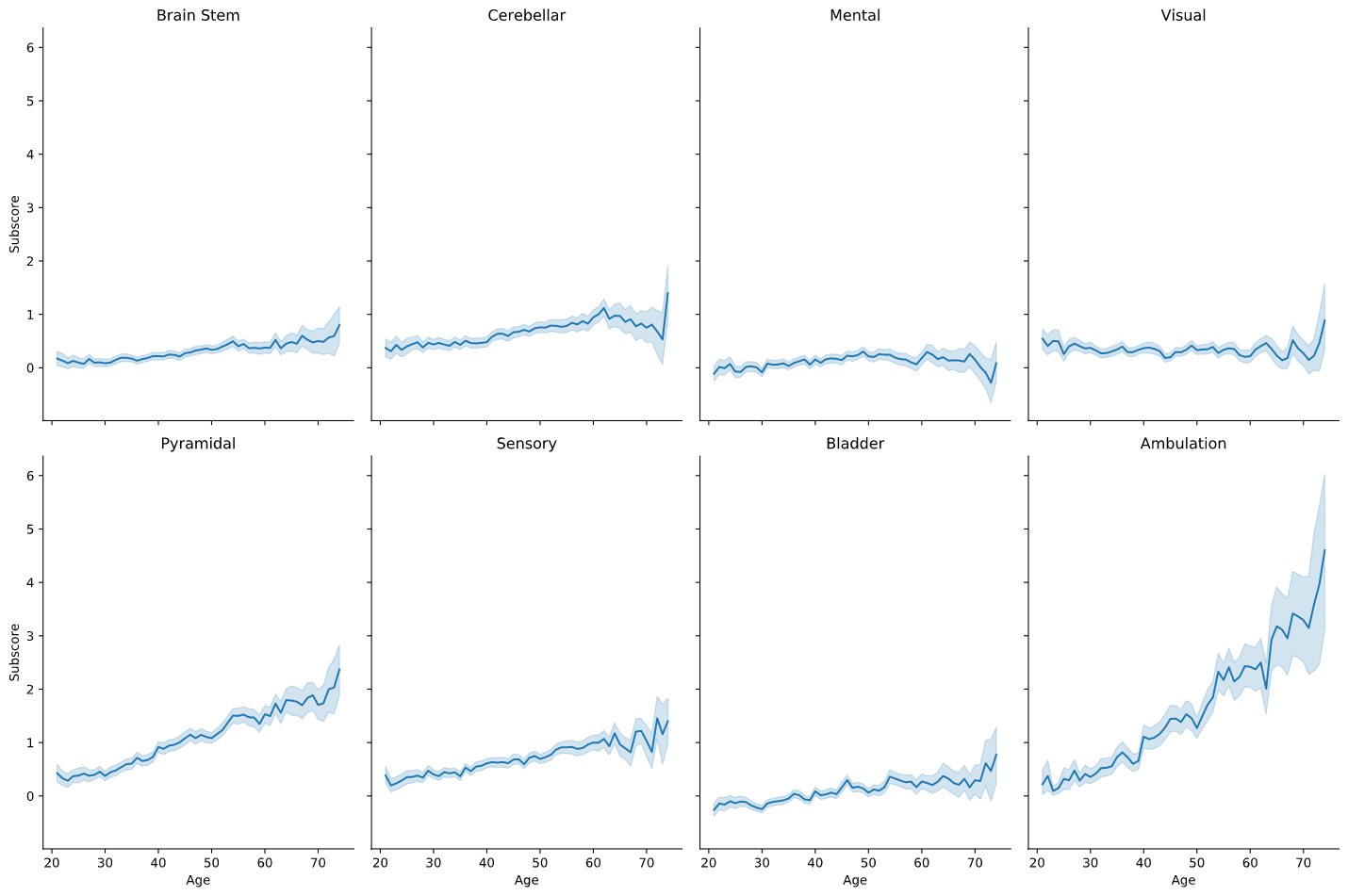}
  \caption{Change of functional subscores with age.}
  \label{figure:edss vs age (3)}
\end{figure*}

\begin{figure*}[htp]
  \centering
  \includegraphics[width=15cm]{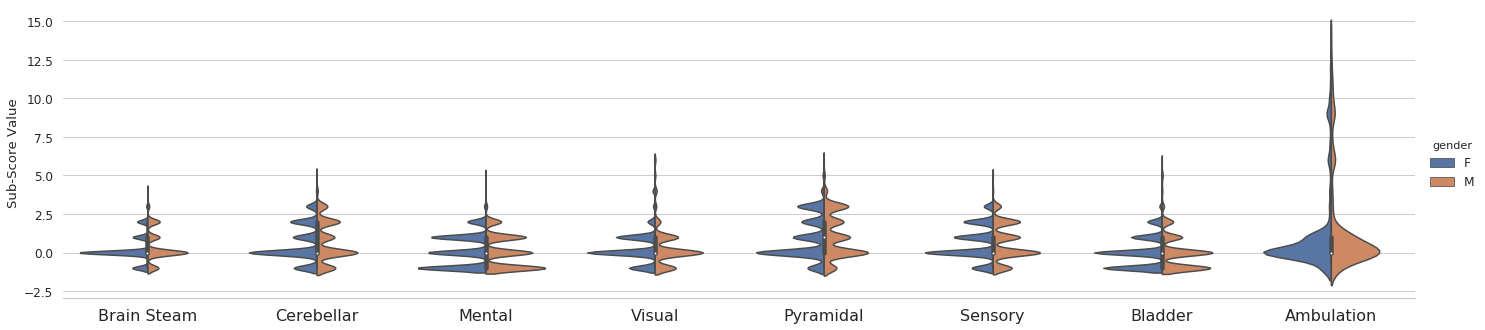}
  \caption{Distribution of functional subscores across gender.}
  \label{figure:edss vs age (4)}
\end{figure*}

\begin{figure*}[htp]
  \centering
  \includegraphics[width=10cm]{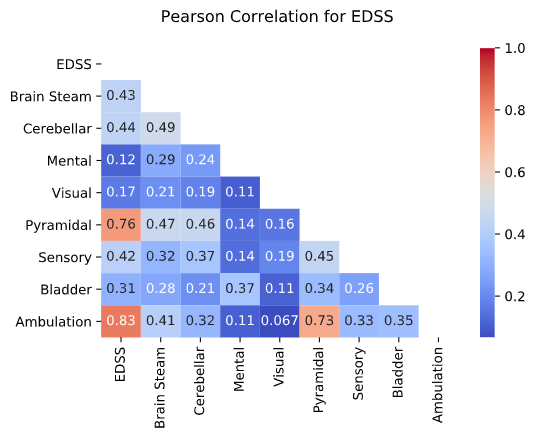}
  \caption{Correlation matrix between functional subscores and EDSS. Strong correlations between EDSS and ambulatory and pyramidal subscores as expected. }
  \label{figure:edss vs age (5)}
\end{figure*}

\end{document}